\documentclass[letterpaper,10pt,conference]{ieeeconf}
\IEEEoverridecommandlockouts
\usepackage{amsmath,amssymb,bm}
\usepackage{booktabs}
\usepackage{graphicx}
\usepackage{multirow}
\usepackage{url}
\usepackage{cite}
\usepackage{xcolor}
\usepackage{etoolbox}
\patchcmd{\abstract}{---\,}{:\ }{}{}
\graphicspath{{figs/}}

\title{\LARGE \bf WorldContact: A Contact-Centric World Model\\
for Scalable Robot Learning}
\author{Caoliwen Wang\textsuperscript{1},
Mengdi Wang\textsuperscript{$^\dagger$,1},
Heng Zhang\textsuperscript{1},\\
Shixun Huang\textsuperscript{1},
Siyuan Chen\textsuperscript{1},
Chao Liu\textsuperscript{1},
Anpei Chen\textsuperscript{2},\\
Zhendong Wang\textsuperscript{3},
Peter Yichen Chen\textsuperscript{$^\dagger$,1},
Huamin Wang\textsuperscript{3}%
\thanks{\textsuperscript{1} University of British Columbia.}%
\thanks{\textsuperscript{2} Westlake University.}%
\thanks{\textsuperscript{3} Style3D Research.}%
\thanks{\textsuperscript{$^\dagger$} Corresponding author.}}

\makeatletter
\providecommand{\bstctlcite}[1]{\@bsphack\if@filesw\immediate\write\@auxout{\string\citation{#1}}\fi\@esphack}
\makeatother

\begin{document}
\bstctlcite{clothparticle_bib_control}
\maketitle
\thispagestyle{empty}
\pagestyle{empty}

\begin{figure*}[t]
\centering
\includegraphics[width=\textwidth]{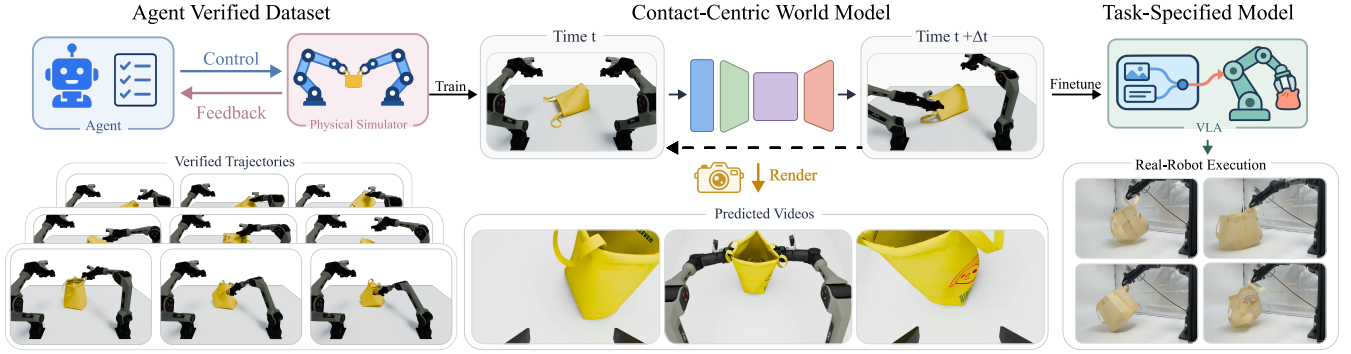}
\caption{\textbf{WorldContact overview.} (Left): an agent-verified physical-simulation dataset, where an automated simulation agent samples the object and prescribed state and executes the resulting controls in the physical simulator. (Middle): contact-centric world model prediction and rendering of object-state trajectories into the observation modality consumed by the VLA. (Right): VLA fine-tuning for task-specific policy adaptation and subsequent real-robot deployment.
}

\label{fig:teaser}
\end{figure*}

\begin{abstract}
Adapting robots to new objects and tasks requires interaction experience
that can be costly to obtain. We present WorldContact, a contact-centric
world model for deformable-object manipulation, constructed from a limited
set of high-quality trajectories to generate additional training data
efficiently. It predicts object
dynamics using larger time steps than the source numerical simulator,
which requires small integration steps to resolve rapid motion and prevent
interpenetration. We evaluate WorldContact across 16 shopping-bag manipulation
tasks. State-rollout measurements on a single H100 GPU show a $10\times$
speedup over the source simulator, excluding rendering and disk I/O. We use the generated data to
fine-tune an existing vision-language-action policy and deploy it directly
on a real robot. In bag lifting, the same policy achieves 65\% single-attempt
success when fine-tuned on source simulation data alone, compared with
95\% when fine-tuned on the dataset expanded with WorldContact. These
results support efficient data generation
with WorldContact for robot policy adaptation.
\end{abstract}

\section{Introduction}

A central goal of embodied AI is to develop robots that can learn and act
across diverse tasks and environments. Household services, industrial
production, and logistics all require adaptation to changing objects,
materials, task requirements, and physical interactions. Learning reusable
manipulation skills from diverse experience can help robots acquire
these capabilities~\cite{rt2,openvla}. To act reliably in these settings,
robots must anticipate how their actions will affect the world. A
\emph{world model} describes this relationship by predicting the outcomes
of actions given the current world state. Such predictive models provide a
basis for reasoning about physical interactions~\cite{worldmodel,dreamer}.

Adapting robot policies to unfamiliar scenes requires relevant interaction
experience. World models can help supply this experience by predicting the
outcomes of robot-object interactions. Constructing suitable world models
is challenging because objects differ in geometry, material properties,
and physical behavior. For deformable-object manipulation, a world model must predict
how objects deform and move in response to robot actions. Yet changing
shapes continually alter physical interactions, from friction between
fabric folds to sliding on support surfaces and robot grasping. Modeling
these interactions across changing configurations makes reliable
prediction difficult~\cite{softgym,softmimicgen,simweaver2026}.

To address this modeling challenge, we introduce \emph{WorldContact},
a contact-centric neural simulator that serves as a world model for
deformable-object manipulation. It builds on the Lagrangian approach of
WorldParticle~\cite{worldparticle} to learn object dynamics from a limited
set of high-quality interaction trajectories and rapidly generate additional
interaction trajectories for robot policy learning. The high-quality source
trajectories may come from physics-based numerical simulation or real-world
interaction. It encodes local geometry and motion around object vertices
to represent interactions within and between objects and with the robot
and environment, enabling prediction of object dynamics under prescribed
robot actions.

To support efficient data generation, WorldContact uses a time step
that is $20\times$ the source simulator's integration step.
Numerical simulation often requires small steps to resolve rapid motion
and prevent interpenetration; WorldContact predicts the resulting state
change over the larger interval directly. Combined with GPU-efficient
inference, this larger time step reduces the cost of generating
policy-training trajectories, with some approximation of high-accuracy
dynamics. In our measurements, WorldContact generates state rollouts
$10\times$ faster than the source simulator, excluding rendering and disk
I/O. We demonstrate this approach on 16 robotic shopping-bag manipulation
tasks, including lifting and supporting a bag. We fine-tune an existing
pretrained vision-language-action (VLA) policy on the expanded simulation
dataset and deploy it directly on a real robot. In the bag-lifting
evaluation, the same VLA succeeds in 13/20 single-attempt trials (65\%)
when fine-tuned only on the 25 source simulation trajectories, compared
with 19/20 trials (95\%) when fine-tuned on the expanded dataset.
These results demonstrate the practical
value of our data-scaling approach: efficiently generating additional
training experience from limited high-quality source data to adapt robot
control policies to new manipulation settings.

Our contributions are:
\begin{itemize}
\clubpenalties 2 10000 0\relax
    \item WorldContact, a contact-centric world model constructed from
    limited high-quality data that encodes local geometry and motion to
    predict object dynamics under robot actions.
    \item A data-generation pipeline that renders predicted trajectories
    and pairs the observations with their conditioning robot actions to
    scale demonstrations for policy fine-tuning.
    \item A shopping-bag study across 16 tasks, demonstrating $10\times$
    faster state-rollout generation and better real-robot bag-lifting
    performance with the expanded dataset than with source data alone.
\end{itemize}


\section{Related Work}

\paragraph{World Models for Robot Learning}
Ha and Schmidhuber~\cite{worldmodel} explored policy learning using latent
world models. Dreamer~\cite{dreamer} learns control policies through imagined
trajectories in a learned latent space. For cloth, VisuoSpatial Foresight
learns RGB-D dynamics in simulation and uses predicted observations to plan
multi-step smoothing and folding~\cite{vsf}. For robotic manipulation,
IRASim generates action-conditioned videos~\cite{irasim}, while Interactive
World Simulator uses learned visual rollouts for policy training and
evaluation, including tasks involving deformable
objects~\cite{interactiveworld}. Together, these works use learned rollouts
for prediction, planning, policy optimization, and demonstration collection.
Their image or latent predictions do not directly
provide the material-point correspondences and boundary-relative geometry
needed for our vertex-error and penetration metrics. Our method maintains
explicit vertex states for these checks and renders observations after
state prediction, separating dynamics validation from observation generation.

\paragraph{Neural Dynamics for Deformable Manipulation}
DPI-Net and GNS represent physical systems as interacting particles and learn
their dynamics through graph message passing~\cite{dpinet,gns}. MeshGraphNets
extends learned simulation to mesh representations, including cloth with
adaptive resolution~\cite{meshgraphnets}. For cloth smoothing, Visible
Connectivity Dynamics infers connectivity between observed points and learns
dynamics on this graph to plan under partial observability~\cite{vcd}.
GraphGarment learns action-conditioned garment dynamics for bimanual
manipulation and uses a residual model for sim-to-real
correction~\cite{graphgarment}. Diffusion Dynamics Models jointly address
full cloth-state reconstruction from partial observations and
action-conditioned prediction for model-predictive folding~\cite{diffusiondynamics}.
RoboCraft learns particle dynamics from RGB-D observations for
elastoplastic shaping~\cite{robocraft}, while RoboCook uses
learned particle dynamics for long-horizon manipulation of elastoplastic
objects with tools~\cite{robocook}. Subsequent work addresses material
adaptation with AdaptiGraph~\cite{adaptigraph}, learning from RGB-D interaction
videos with Particle-Grid Neural Dynamics~\cite{particlegrid}, and multiple
objects and materials with ParticleFormer~\cite{particleformer}. PointWorld
represents scene motion and robot actions as 3D point flows to learn
manipulation dynamics across embodiments~\cite{pointworld}. SoMA learns
robot-conditioned soft-body dynamics over reconstructed 3D Gaussian splats
from real observations~\cite{soma}. Contact-Aware Neural Dynamics uses tactile
contact information to improve forward prediction and sim-to-real policy
alignment in rigid-object grasping~\cite{contactaware}. These approaches motivate explicit interaction
modeling, while differing in supervision, state representation, and access to
contact measurements. Our method uses full vertex-state supervision from
a calibrated teacher and explicitly encodes spatial, topological, controllable,
and environmental contact.

WorldParticle models Lagrangian particle dynamics with a transformer that
combines local interaction encoding and global
communication~\cite{worldparticle}. WorldContact adopts
this architectural basis to condition object dynamics on
prescribed robot motion and environmental geometry. Our contact-centric
formulation models interactions among objects, the robot, and the
environment, including contact through rigidly held objects.
We evaluate the resulting predictions through dynamics fidelity and
their use in training a real-robot policy.

\paragraph{Simulation and Sim-to-Real for Deformable Manipulation}
Visual domain randomization addresses observation differences~\cite{tobin2017domain},
while simulation adaptation uses real rollouts to adjust the distribution of
simulated dynamics~\cite{chebotar2019closing}. SoftGym provides deformable
manipulation tasks for robot learning~\cite{softgym}; GarmentLab and
DexGarmentLab expand garment assets, interactions, and manipulation
benchmarks~\cite{garmentlab,dexgarmentlab}. DexGarmentLab also uses garment
correspondences to generate trajectories from a single expert demonstration.
FlingBot learns dynamic cloth
unfolding~\cite{flingbot}, while Cloth Funnels combines dynamic and
quasi-static actions to canonicalize and align garments for downstream
tasks~\cite{clothfunnels}. Alongside these environments and policies,
MimicGen expands demonstrations by transforming and executing source
trajectories~\cite{mimicgen}, and SoftMimicGen extends demonstration generation
to deformable objects~\cite{softmimicgen}. SimWeaver combines measurement-backed
deformable simulation, trajectory synthesis, and sensor-aware variation in
rendered observations for RGB sim-to-real transfer without per-task
calibration~\cite{simweaver2026}. These systems
address the availability and diversity of simulated experience. Our work uses a learned world model to reuse existing state trajectories
for further interaction-data generation. In the simulation-seeded
implementation evaluated here, calibrated physical rollouts supply
supervision, and model rollouts generate additional trajectories. Teacher
calibration and coverage therefore remain relevant to this setting.

At the solver and asset level, DiffCloth differentiates through dry frictional
contact for cloth parameter estimation and control~\cite{diffcloth}.
DiffCP identifies cloth material parameters from RGB-D observations and uses
the calibrated model for state estimation~\cite{diffcp}. PhysTwin reconstructs
deformable digital twins from sparse interaction videos, combining spring-mass
dynamics with Gaussian-splat rendering~\cite{phystwin}. Our evaluated implementation uses
calibrated assets and studies the learned transition used for subsequent
rollout generation.

\section{Method}
\label{sec:method}

We represent the scene using two complementary components: the object state, which jointly describes all objects whose dynamics are to be predicted, and the prescribed state, which specifies robot motion and external geometry supplied as inputs. At step $t$, the object state is represented as
\begin{equation}
\mathcal S_t=[\bm X_t,\bm V_t, \mathcal E, \bm C],
\label{eq:target_state}
\end{equation}
where $\bm X_t,\bm V_t\in\mathbb R^{N\times3}$ denote the vertex positions and velocities, respectively, and $\mathcal E$ represents the rest topology of the target objects. For vertex i, $c_i$ concatenates an object-identity tag with its
rest-state (u,v) coordinates, which specify its location in the object's
topological space. The attributes of all predicted vertices are stacked to
form $\bm C\in\mathbb R^{N\times3}$.

The prescribed state is represented as
\begin{equation}
\mathcal B_t=[\bm a_t,\bm G,\bm X_t^{\rm E},\bm V_t^{\rm E},
\bm C_t^{\rm E}],
\label{eq:boundary_state}
\end{equation}
where $\bm a_t=[\bm q_t^{\rm L},d_t^{\rm L},
\bm q_t^{\rm R},d_t^{\rm R}]\in\mathbb R^{14}$,
$\bm q_t^{\rm L},\bm q_t^{\rm R}\in\mathbb R^6$ are commanded joint rotations
for the two arms, and $d_t^{\rm L},d_t^{\rm R}$ are their gripper openings.
$\bm G$ specifies the fixed kinematic structure and the geometries of its links, gripper, and objects rigidly held by the robot.
$\bm X_t^{\rm E}\in\mathbb R^{N_t^{\rm E}\times3}$ and velocities
$\bm V_t^{\rm E}\in\mathbb R^{N_t^{\rm E}\times3}$ represent positions and velocities of sampled environmental geometry, such as the table. 
The environmental attributes
$\bm C_t^{\rm E}\in\mathbb R^{N_t^{\rm E}\times4}$ concatenate
environment tags and surface
normals. Each environment
tag identifies the geometry to which the sample point belongs. 
An object is included in $\mathcal{S}_t$ if its state is predicted by the world model, and in $\mathcal{B}_t$ if its state is prescribed externally. For example, a ball released by the robot is included in $\mathcal{S}_t$ because its subsequent motion must be predicted, whereas a ball rigidly held by the robot is included in $\mathcal{B}_t$ because its motion is prescribed by the robot.

\begin{figure*}[t]
    \centering
    \includegraphics[width=\linewidth]{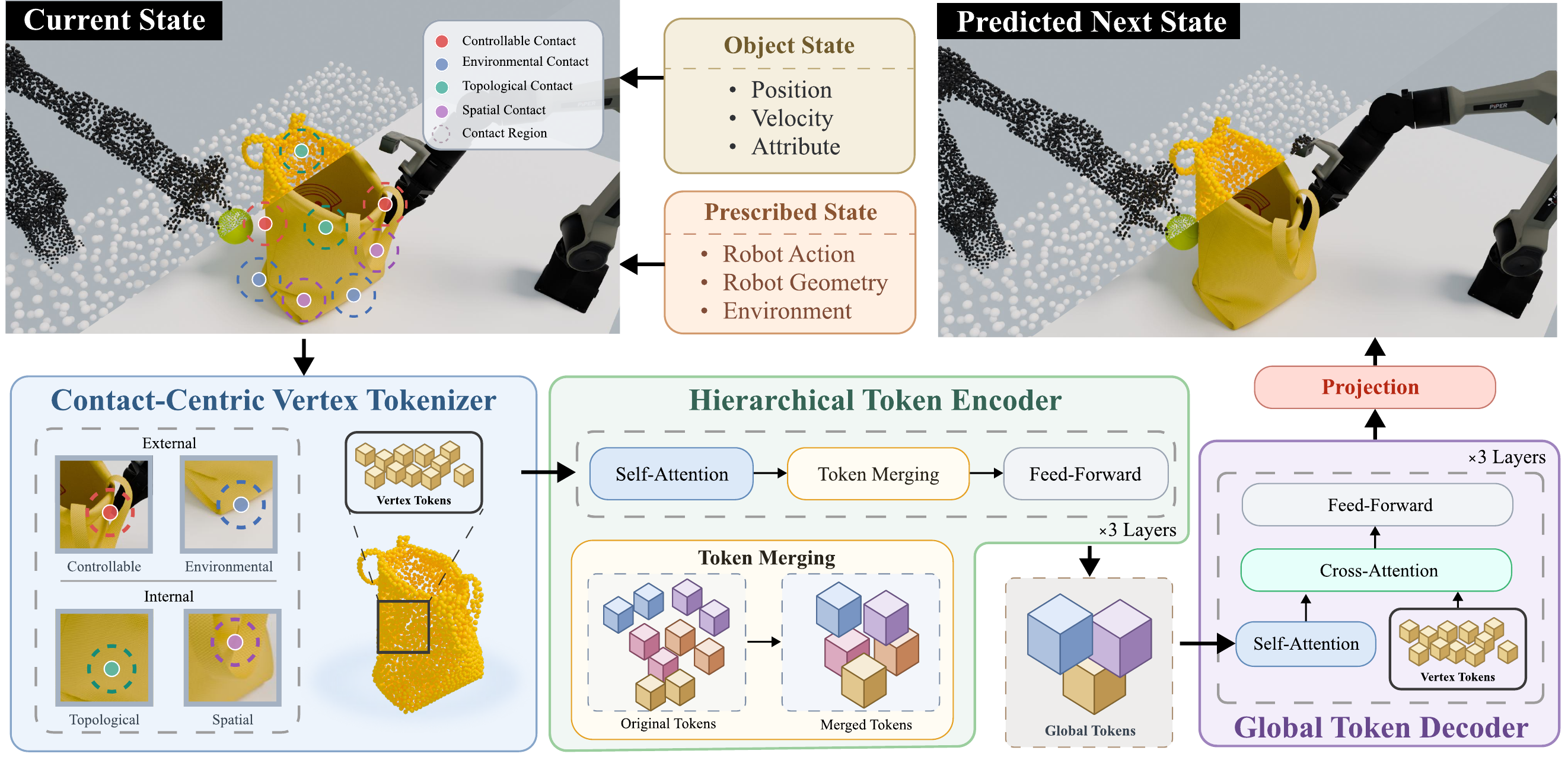}
    \caption{\textbf{WorldContact pipeline.} Given the current object state and prescribed state, WorldContact proceeds through four stages: the contact-centric vertex tokenizer encodes spatial, topological, controllable, and environmental contact information into vertex tokens; the hierarchical encoder merges vertex tokens into global tokens for long-range communication; and the global-token decoder refines vertex tokens to produce vertex-level state corrections. A differentiable projection is used to improve long-horizon rollout stability during fine-tuning.
}
    \label{fig:pipeline}
\end{figure*}

Our model learns how the object state evolves under contact with the prescribed state:
\begin{equation}
(\Delta\bm X_t,\Delta\bm V_t)
=\mathcal F_\theta(\mathcal S_t,\mathcal B_{t+\Delta t}).
\label{eq:student_corrector}
\end{equation}
Then we update the target next state as
$\widehat{\bm X}_{t+\Delta t}=\bm X_t+\Delta\bm X_t$ and
$\widehat{\bm V}_{t+\Delta t}=\bm V_t+\Delta\bm V_t$. Figure~\ref{fig:pipeline} shows the model architecture.

\subsection{Contact-Centric Vertex Tokenizer}
For each target vertex $i$, we encode the local geometry and motion associated with the following contacts:

\begin{itemize}
    \item \textit{Spatial contact.} $\mathcal{N}^{\rm S}_{i}$ contains vertices that are close to $i$ in Euclidean space across all target objects. It captures contacts between different parts or objects, such as between a bag's strap and body or between two predicted objects.

    \item \textit{Topological contact.} $\mathcal{N}^{\rm T}_{i}$ contains vertices that are nearby in the rest-state topology $\mathcal E$ of the same object, represented in UV space. It captures local interactions between vertices connected through the object's intrinsic topology.

    \item \textit{Controllable contact.} $\mathcal{N}^{\rm C}_{i}$ contains nearby samples from robot links, grippers, and objects rigidly held by the robot. It captures direct robot--object contact as well as contact mediated by held objects, such as a rigid tool pushing the bag.

    \item \textit{Environmental contact.} $\mathcal{N}^{\rm E}_{i}$ contains nearby samples from environmental geometry, such as a table supporting the bag.
\end{itemize}
We denote the positions, velocities, and attributes of controllable samples at time t by $\bm X^{\rm C}_{t},\bm V^{\rm C}_{t},\bm C^{\rm C}_{t}$. Controllable samples follow the commanded configuration $\bm a_{t}$ and kinematic geometry $\bm G$. Their attributes $\bm C^{\rm C}_{t}$ use the tags and surface
normals as $\bm C^{\rm E}_{t}$. 
For $k\in\{\rm S,T,C,E\}$, a compact-support learned
kernel aggregates neighbor features:
\begin{equation}
\bm g_i^k=\sum_{j\in\mathcal N_i^k}
\bm W_k(\bm r_{ij}^k)\bm u_{ij}^k.
\label{eq:contact_aggregation}
\end{equation}
The corresponding displacements and features are
\begin{equation}
\resizebox{0.90\columnwidth}{!}{$
\begin{aligned}
\textit{Spatial Contact:}\quad
&\bm r_{ij}^{\rm S}=\bm x_{t,j}-\bm x_{t,i},
&&\bm u_{ij}^{\rm S}=[\bm v_{t,j},\bm c_j],\\
\textit{Topology Contact:}\quad
&\bm r_{ij}^{\rm T}=\bm x_{t,j}-\bm x_{t,i},
&&\bm u_{ij}^{\rm T}=[\bm v_{t,j},\bm c_j],\\
\textit{Controllable Contact:}\quad
&\bm r_{ij}^{\rm C}=\bm x_{t+\Delta t,j}^{\rm C}-\bm x_{t,i},
&&\bm u_{ij}^{\rm C}=[\bm v_{t+\Delta t,j}^{\rm C},\bm c_{t+\Delta t,j}^{\rm C}],\\
\textit{Environmental Contact:}\quad
&\bm r_{ij}^{\rm E}=\bm x_{t+\Delta t,j}^{\rm E}-\bm x_{t,i},
&&\bm u_{ij}^{\rm E}=[\bm v_{t+\Delta t,j}^{\rm E},\bm c_{t+\Delta t,j}^{\rm E}].
\end{aligned}
$}
\label{eq:contact_branches}
\end{equation}
The resulting vertex token is
\begin{equation}
\bm h_i=\operatorname{MLP}
([\bm x_{t,i},\bm v_{t,i},\bm c_i,
  \bm g_i^{\rm S},\bm g_i^{\rm T},\bm g_i^{\rm E},
  \bm g_i^{\rm C}]).
\label{eq:tokenizer}
\end{equation}
Each token thus combines the target vertex's state and attributes with local geometric features and contact information.

\subsection{Hierarchical Token Encoder}
The vertex tokens ${\bm h_i}$ effectively encode local geometric information and external contact interactions. However, many physical phenomena in deformable objects exhibit long-range coupling. For example, when a cloth is stretched by pulling one region, vertices that are spatially distant from each other may undergo similar motions because they experience similar deformation and force patterns. To model such long-range coupling, self-attention provides a natural mechanism for propagating information across distant vertices~\cite{vaswani2017attention}. Applying self-attention repeatedly over all $N$ vertices, however, is computationally expensive. We adopt WorldParticle's hierarchical token-merging strategy~\cite{worldparticle} to reduce the number of tokens involved in global attention.

Let $\bm H_{\rm e}^{(0)}=[\bm h_i]_{i=1}^{N}$,
$\bm X^{(0)}=\bm X_t$, and $m_i^{(0)}=1$. 
Each encoder level first exchanges
global context at its current resolution using multi-head self-attention~\cite{vaswani2017attention}
with 3D rotary positional encoding (RoPE)~\cite{worldparticle}:
\begin{equation}
\widehat{\bm H}_{\rm e}^{(\ell)}
=\operatorname{SelfAttn}
(\bm H_{\rm e}^{(\ell-1)},\bm X^{(\ell-1)}).
\label{eq:encoder_attention}
\end{equation}
Tokens are then divided into two alternating partitions. Each token in the first is
matched to the token in the second with the highest cosine similarity, reducing
the count to
$N^{(\ell)}=\lceil N^{(\ell-1)}/2\rceil$. If
$\mathcal M_i^{(\ell)}$ is the set assigned to output token $i$, the merged
feature, anchor, and multiplicity are
\begin{align}
m_i^{(\ell)}&=\sum_{j\in\mathcal M_i^{(\ell)}}m_j^{(\ell-1)},\nonumber\\
(\bm h_i^{(\ell)},\bm x_i^{(\ell)})
&=\sum_{j\in\mathcal M_i^{(\ell)}}
\frac{m_j^{(\ell-1)}}{m_i^{(\ell)}}
(\widehat{\bm h}_j^{(\ell)},\bm x_j^{(\ell-1)}).
\label{eq:merge}
\end{align}
Repeated merging yields
$(\overline{\bm H},\overline{\bm X})
=(\bm H_{\rm e}^{(L_e)},\bm X^{(L_e)})$ with
$N_g\ll N$ global tokens. 
Each global token summarizes the local geometric, contact, velocity, and attribute information of the vertex group it represents. Together, these tokens form a compact representation of the current object state that supports efficient information sharing across spatially distant regions.

\subsection{Global Token Decoder}
The original vertex tokens bypass compression and initialize the queries
$\bm H_{\rm d}^{(0)}=\bm H_{\rm e}^{(0)}$; the global tokens start from
$\bm H_{\rm g}^{(0)}=\overline{\bm H}$. 
Subsequent self-attention operates on these global tokens to capture long-range dependencies:
\begin{equation}
\bm H_{\rm g}^{(\ell)}
=\operatorname{SelfAttn}
(\bm H_{\rm g}^{(\ell-1)},\overline{\bm X}).
\label{eq:decoder_self}
\end{equation}
The updated global representation is then queried by the vertex-resolution tokens through cross-attention:
\begin{equation}
\bm H_{\rm d}^{(\ell)}
=\operatorname{CrossAttn}
(\bm H_{\rm d}^{(\ell-1)},\bm X_t;
 \bm H_{\rm g}^{(\ell)},\overline{\bm X}),
\label{eq:decoder_cross}
\end{equation}
which allows each vertex query to retrieve global context relevant to its local state while retaining the local geometric and contact information encoded in the vertex tokens. A pointwise output head predicts the corrections:
\begin{equation}
(\Delta\bm x_{t,i},\Delta\bm v_{t,i})
=\operatorname{MLP}_{\rm out}(\bm h_{{\rm d},i}^{(L_d)}).
\label{eq:state_head}
\end{equation}

\subsection{Training}
\label{sec:model_training}
We learn WorldContact from high-quality interaction trajectories, using their recorded states as supervision. During training, we sample trajectory segments of length $W$ and autoregressively roll out the model over the segment. The training objective is computed over the predicted states at all rollout steps, averaging the position and velocity errors across all predicted vertices:
\begin{align}
\mathcal L
&=\frac{1}{N(W-1)}\sum_{n=1}^{W-1}\sum_{i=1}^{N}
\left(\lambda_x\|\widehat{\bm x}_{t+n \Delta t,i}-\bm x_{t+n \Delta t,i}\|_2^2\right.\nonumber\\
&\hspace{31mm}\left.+\lambda_v\|\widehat{\bm v}_{t+n \Delta t,i}-\bm v_{t+n \Delta t,i}\|_2^2\right).
\label{eq:loss}
\end{align}
Gradients propagate through all rollout steps to reduce error accumulation during autoregressive prediction.

Starting from the base model, we continue training with differentiable topology projection $\mathcal P$ inside the autoregressive prediction loop to improve long-horizon stability. Let $\widetilde{\bm X}_{t+\Delta t}$ denote the unconstrained position predicted by the network. The state used by the rollout is
\begin{equation}
\widehat{\bm X}_{t+\Delta t}=\mathcal P(\widetilde{\bm X}_{t+\Delta t},\mathcal E),\qquad
\widehat{\bm V}_{t+\Delta t}=\frac{\widehat{\bm X}_{t+\Delta t}-\widehat{\bm X}_{t}}{\Delta t}.
\label{eq:projection_finetune}
\end{equation}
The projection smooths the predicted displacement field and limits topology-edge distortion. Its output is used in Eq.~\eqref{eq:loss} and fed into the next autoregressive step, allowing gradients to propagate through both the projection iterations and the full rollout.

\section{Experiments}
\label{sec:experiments}

\subsection{Data Collection}
\begin{figure}[t]
    \centering
    \includegraphics[width=0.48\textwidth]{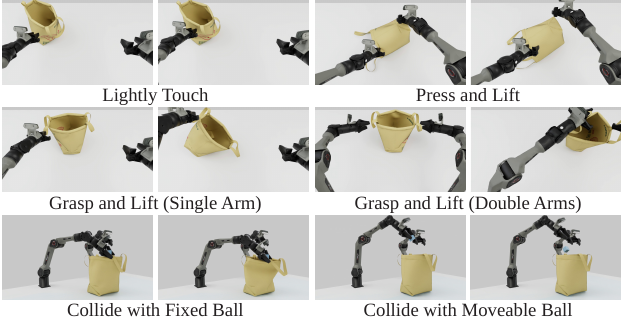}
    \vspace*{-1em}
    \caption{\textbf{Action and contact diversity in the collected corpus.}
Each panel shows a representative shopping-bag task, including object collision, grasping, lifting, placing, and releasing, covering complex contacts.
}
    \label{fig:dataset_action_diversity}
\end{figure}

We develop an automated simulation agent powered by GPT-5.6 Sol with
xhigh reasoning effort to configure and run tasks in the physical simulator
used by SimWeaver~\cite{simweaver2026}. The agent monitors the resulting
trajectories and discards executions that fail the checks described below.
To align simulated dynamics with real-world behavior, we first reconstructed
each target object as a digital asset and calibrated its simulation
parameters using real-world observations. We discretize the target bag into
20,274 vertices, while rigid spheres and cubes use 951 and 465 vertices,
respectively. The robot gripper and table are represented by 1,114 and
16,900 sampled points, respectively.
Simulator parameters and solver settings are fixed across the dataset, with a 0.002s physical time step chosen to ensure numerical stability; object-specific parameters, including friction, damping, and Young's modulus, are fixed for each object to be predicted.
For each execution, the agent first samples the initial object state. It then partitions feasible object configurations and interaction locations into coverage cells, visits these cells while varying the interaction speed and motion amplitude, and records the resulting trajectory together with its conditioning action.
The dataset covers 16 shopping-bag manipulation and collision tasks, including object collision, grasping, lifting, placing, and releasing. Together, these tasks involve spatial, topological, controllable, and environmental contacts, including simultaneous interactions, with a subset shown in Fig.~\ref{fig:dataset_action_diversity}.

Each trajectory passes online physical and visual checks. Physical checks reject
states containing undefined or infinite values (NaN or Inf), solver failures,
interpenetration, self-intersection, and
spurious gripper--bag adhesion. Visual checks reject executions that move the
bag outside the workspace or camera view. The state machine is
\textsc{Reset} $\rightarrow$ \textsc{Execute} $\rightarrow$ \textsc{Validate}
$\rightarrow$ \textsc{Accept}; a failed check terminates the rollout and logs
its first failing predicate. Only trajectories that pass every check are used
for training.

The agent maintains a memory of past execution outcomes and failure cases, allowing it to learn from previous failures and avoid repeating similar failures in future executions.

\subsection{Training Details}
We use the same dataset split for initial WorldContact training and
subsequent training with differentiable topology projection inside the
prediction loop (Sec.~\ref{sec:model_training}). The dataset covers 16
shopping-bag manipulation tasks, with 25 training trajectories per task
(400 in total), including two validation and two test trajectories per task.
Both training stages use 8 NVIDIA H100 GPUs, five-step rollout
segments with a time step of 0.04 s, and equal position and velocity loss
weights of $10^4$. Further training details and model specifications are
provided in the Appendix.

\subsection{Neural Dynamics Comparison}
\label{sec:baseline_comparison}
\begin{figure*}[t]
    \centering
    \includegraphics[width=\textwidth]{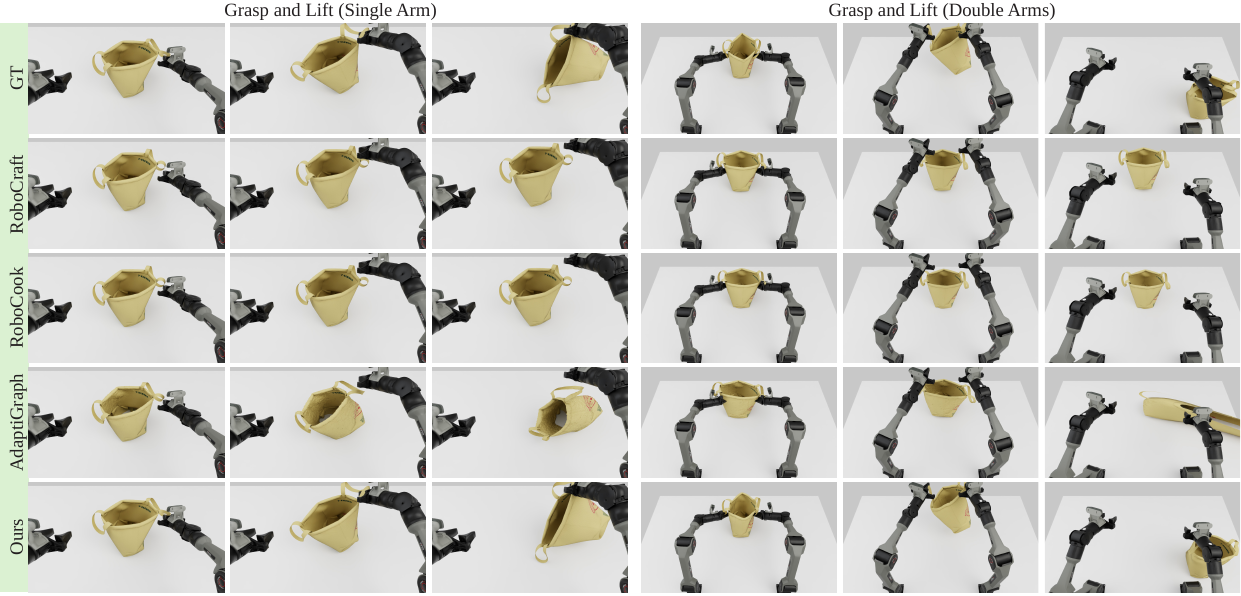}
    \vspace*{-1em}
    \caption{\textbf{Qualitative rollout comparison.} The top row shows the ground-truth states, while the subsequent rows show rollouts from RoboCraft~\cite{robocraft}, RoboCook~\cite{robocook}, AdaptiGraph~\cite{adaptigraph}, and Ours. Columns correspond to successive time steps. RoboCraft~\cite{robocraft} and RoboCook~\cite{robocook} fail to maintain the bag–gripper contact during the lifting process, causing the bag to remain stationary. AdaptiGraph suffers from severe physically implausible deformation. In contrast, our method successfully simulates the lifting process while maintaining stable rollouts. The rendered views on the left are consistent with those used for VLA fine-tuning.
}
    \label{fig:contact_rollouts}
\end{figure*}

We compare WorldContact with three neural dynamics baselines: RoboCraft~\cite{robocraft},
RoboCook~\cite{robocook}, and AdaptiGraph~\cite{adaptigraph}, shown in Fig.~\ref{fig:contact_rollouts}.
All methods are trained on the same dataset with identical dataset splits. For autoregressive test rollouts, we report cloth
vertex position and velocity mean squared error (MSE) against ground truth,
averaged over all vertices and frames. The penetration rate ($\%$) measures the percentage of cloth contact area penetrating the collision geometry by more than \(1\,\mathrm{cm}\), weighted by vertex-associated mesh areas and aggregated across all frames. Table~\ref{tab:main_comparison}
reports the aggregate test results; lower values are better for all metrics.

\begin{table}[t]
\centering
\scriptsize
\setlength{\tabcolsep}{2.4pt}
\caption{Neural dynamics comparison.}
\label{tab:main_comparison}
\begin{tabular}{@{}lccc@{}}
\toprule
Method & Position MSE $\downarrow$ & Velocity MSE $\downarrow$
& Penetration Rate (\%) $\downarrow$\\
\midrule
RoboCraft~\cite{robocraft}
& 0.0598 & 0.0106 & 13.88\%\\
RoboCook~\cite{robocook}
& 0.0575 & 0.0105 & 10.28\%\\
AdaptiGraph~\cite{adaptigraph}
& 0.0323 & 0.0098 & 24.84\%\\
WorldContact
& \textbf{0.0002}
& \textbf{0.0002}
& \textbf{0.35\%}\\
\bottomrule
\end{tabular}
\end{table}

\subsection{Computational Efficiency}
\label{sec:computational_efficiency}
To evaluate computational efficiency, we compare complete state rollouts generated by WorldContact and the physical simulator used for data generation, using the same initial object state and prescribed-state inputs. All measurements are performed on a single NVIDIA H100 GPU, excluding rendering and disk I/O. WorldContact simulates 0.4 s of dynamics per second of wall-clock time, compared with 0.04 s for the physical simulator, yielding a 10$\times$ speedup.

\subsection{Ablation Study}
\label{sec:ablations}

\begin{figure}[t]
    \centering \includegraphics[width=0.48\textwidth]{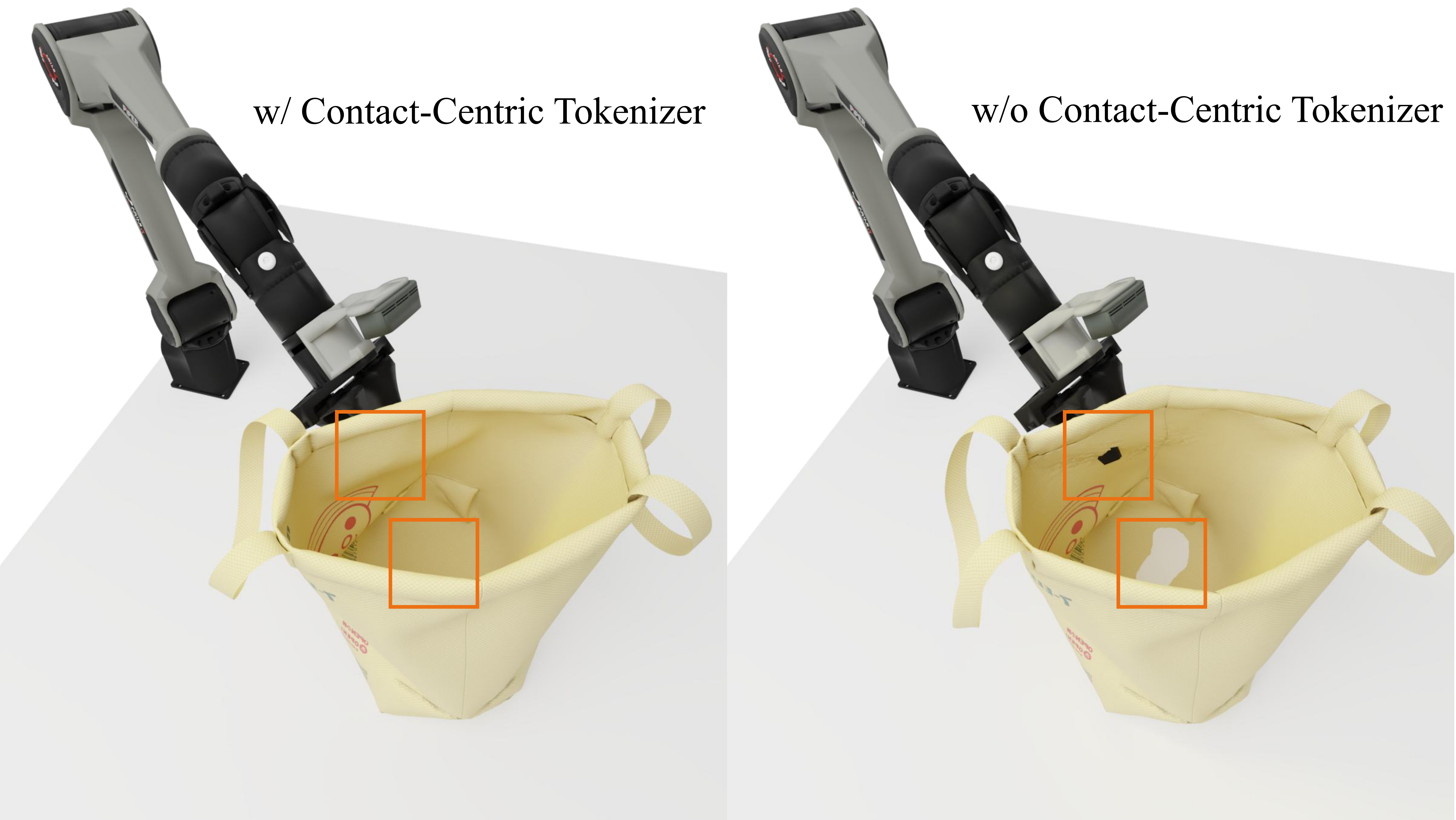}
    \vspace*{-1em}
    \caption{\textbf{Ablation of the contact-centric vertex tokenizer.} Same state with (left) and without (right) the contact-centric vertex tokenizer. Orange boxes highlight the corresponding contact regions, where removing our tokenizer leads to penetration.}
\label{fig:contact_tokenizer_ablation}
\end{figure}

\paragraph{Contact-centric vertex tokenizer}
We ablate the contact-centric vertex tokenizer. For each vertex $i$, we
remove the spatial, topological, controllable, and environmental contact features
$\bm g_i^{\rm S}$, $\bm g_i^{\rm T}$, $\bm g_i^{\rm C}$, and
$\bm g_i^{\rm E}$, and form its token by linearly projecting the concatenation of $[\bm{x}_{t,i}, \bm{v}_{t,i}, \bm{c}_i]$ with the features of its nearest environmental and controllable sample into the same dimension as $\bm{h}_i$.
The downstream architecture remains unchanged. 
We train the new architecture from scratch using exactly the same training strategy throughout. As shown in Fig.~\ref{fig:contact_tokenizer_ablation}, removing these contact features leads to vertex penetration into the table and gripper, resulting in unstable rollouts and contact failures, while using our tokenizer maintains stable contacts.

\paragraph{Hierarchical token compression}
We ablate the hierarchical token compression paradigm of our method. Specifically,
we replace the hierarchical token encoder and global token decoder with full-resolution
self-attention blocks with the same number of attention layers. We report the estimated training time per optimization step for training segment lengths $W=5,10,15$ on a single H100 GPU. Table~\ref{tab:global_token_ablation}
shows that global-token communication reduces the time per optimization
step while preserving the vertex-level output; lower values are better.

\begin{table}[t]
\centering
\scriptsize
\setlength{\tabcolsep}{5pt}
\caption{Training time per optimization step (s).}
\label{tab:global_token_ablation}
\begin{tabular}{@{}c ccc@{}}
\toprule
Attention Type
& $W=5$ $\downarrow$
& $W=10$ $\downarrow$
& $W=15$ $\downarrow$ \\
\midrule
w/o hierarchical token compression
& 4.367
& 11.236
& 15.873 \\

w/ hierarchical token compression
& \textbf{4.184}
& \textbf{10.526}
& \textbf{14.493} \\
\bottomrule
\end{tabular}
\end{table}

\section{Application}
\label{sec:application}

We fine-tune an existing VLA policy for single-arm bag lifting using a larger dataset produced with WorldContact from the initial high-quality simulation data. Given an initial object state and a sequence of prescribed-state inputs, the world model predicts an object-state trajectory, which is then rendered into the observation modality consumed by the VLA~\cite{pi05}. Each rendered observation is paired with the corresponding control command used to condition the rollout, ensuring consistent observation-action alignment. For this evaluation, we use 25 high-quality simulated bag-lifting trajectories as source data and set the expanded policy-training dataset size to 150 trajectories. We adopt a right-arm-only control interface and provide the policy with two RGB views captured by a top-mounted camera and a right-wrist-mounted camera. The language instruction is ``right grasp and lift.'' Starting from the pretrained Pi05 checkpoint, we fine-tune the policy for the target task. Details are provided in Appendix.

After fine-tuning, we freeze the policy and deploy it directly on the real robot. The task requires a single attempt to grasp the bag, lift it above the table, and hold it for 2 seconds without human intervention. The same VLA fine-tuned only on the 25 source trajectories succeeds in 13 of 20 trials (65\%). Fine-tuning with the expanded dataset increases success to 19 of 20 trials (95\%). This task-level comparison supports the usefulness of the generated data for policy adaptation.

\begin{figure}[t]
    \centering
    \includegraphics[width=0.48\textwidth]{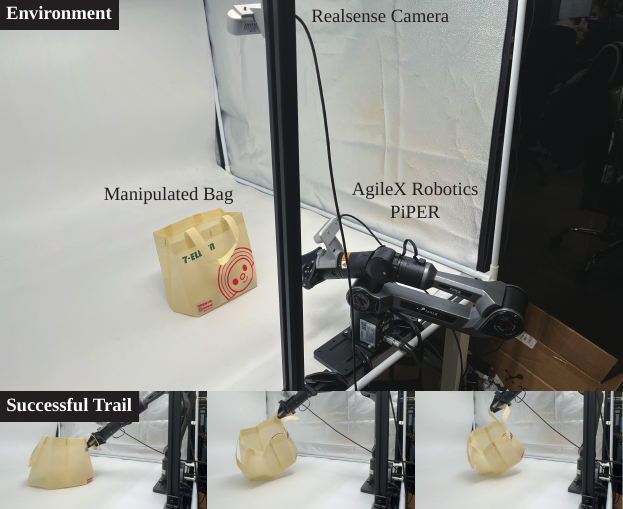}
    \vspace*{-1em}
    \caption{\textbf{Real world experiment.} The top row shows our experimental setup, including the AgileX PiPER 6-DoF robotic arm, the manipulated 7-Eleven bag, and two RealSense cameras: a D435i top-mounted camera and a D435 right-wrist-mounted camera. The
    bottom row shows a successful trial.
    }
    \label{fig:vla}
\end{figure}

Figure~\ref{fig:vla} illustrates our real-world experimental setup, including the robot platform, the bag used as the target object, and the camera configuration. It also shows a representative successful execution sequence, demonstrating the complete process from approaching and grasping the bag to lifting it above the table and maintaining the lifted state.

\section{Conclusion}
\label{sec:conclusion}

We presented WorldContact, a contact-centric world model constructed from
limited high-quality interaction data to generate training experience
efficiently for robot learning. Its larger time step and GPU-efficient
inference enable state-rollout generation at $10\times$ the speed of the
source numerical simulator. We evaluated
WorldContact across 16 shopping-bag manipulation tasks and used its
generated data for VLA fine-tuning. For real-robot bag lifting, the same
VLA achieved 65\% single-attempt success when fine-tuned on source simulation
data alone, compared with 95\% when fine-tuned on the dataset expanded with
WorldContact. These results support efficient data generation for robot policy
adaptation.

Prediction errors under distribution shift and differences between rendered
and real observations remain challenges. Future work will examine how
source-data coverage, prediction fidelity, and generated-data budgets
affect downstream policy performance. Policy failures could guide the
collection of additional real or simulated interactions and subsequent
world-model updates. Building WorldContact from real-world interaction data
and evaluating transfer to unseen objects, materials, and tasks will
further test its potential for scalable robot learning.

\appendix
\paragraph{WorldContact Specification}
The particle tokenizer uses \(4\times4\times4\) continuous-convolution kernels with width 64. Fluid and topology features each have 32 channels, while boundary and direct particle features each have 64 channels, yielding \(\mathbf{h}_i\in\mathbb{R}^{192}\). The super-token encoder performs \(L_e=3\) self-attention and bipartite token-merging iterations with 8 heads (head dimension 24, \(k=2\)) and 3D RoPE of dimension 24. The decoder contains \(L_d=3\) cross/self-attention layers at width 192, with 8 heads, FFN width 128, and dropout 0.1. The prediction head is a \(192\rightarrow128\rightarrow128\rightarrow6\) MLP. The tokenizer, super-token encoder, decoder, and prediction head contain 57,993, 782,784, 1,116,480, and 42,502 trainable parameters, respectively, for a total of 1,999,759 parameters (approximately 2.00M).

\paragraph{WorldContact Training Details} We use AdamW with $\boldsymbol{\beta}=(0.9,0.999)$ and a weight decay of
$5\times10^{-4}$. Base training uses a peak learning rate of
$1\times10^{-4}$, with an 8,000-step linear warmup from $1\times10^{-6}$
followed by cosine decay toward $5\times10^{-6}$, and runs for 93,240
optimization steps. Starting from the pretrained model, continued training
with differentiable topology projection uses a peak learning rate of
$2\times10^{-5}$, with an 8,000-step linear warmup from $2\times10^{-7}$
followed by cosine decay toward $5\times10^{-6}$, and runs for 31,080
optimization steps. After each predicted step, we apply a differentiable
projection with four Laplacian-smoothing iterations (weight 0.35), followed
by 32 edge-constraint iterations that restrict each edge length to
$0.9$--$1.1$ times its rest-topology length.

\paragraph{Policy Fine-Tuning Details} We use AdamW with $\beta=(0.9,0.95)$, $\epsilon=1\times10^{-8}$, and weight decay $1\times10^{-2}$. Training uses an initial learning rate of $2.5\times10^{-5}$, with a 1,000-step linear warmup followed by cosine decay to $2.5\times10^{-6}$ over a total of 30,000 optimization steps. State and action features are normalized using quantile normalization, while visual features use identity normalization. The policy predicts 50 future action steps per chunk, and we perform full-model fine-tuning starting from the pretrained checkpoint.

\bibliographystyle{IEEEtran}
\bibliography{references}
\end{document}